\documentclass[conference]{IEEEtran}
\ifCLASSINFOpdf
  \usepackage[pdftex]{graphicx}
\else
\fi
\usepackage{subfig}
\usepackage[nocompress]{cite}

\usepackage{amsmath}
\usepackage{amsfonts}
\usepackage{mathtools,xparse}
\usepackage[section]{placeins}

\DeclarePairedDelimiter{\norm}{\lVert}{\rVert}
\NewDocumentCommand{\normL}{ s O{} m }{%
  \IfBooleanTF{#1}{\norm*{#3}}{\norm[#2]{#3}}_{L_2(\Omega)}%
}

\begin{document}
%
\title{Rethinking Convolutional Semantic Segmentation Learning}

\author{\IEEEauthorblockN{Mrinal Haloi}
\IEEEauthorblockA{IIT Guwahati, ARTELUS\\
Email: h.mrinal@iitg.ernet.in}
}


%


\maketitle

\begin{abstract}
Deep convolutional semantic segmentation (DCSS) learning doesn't converge to an optimal local minimum with random parameters initializations; a pre-trained model on the same domain becomes necessary to achieve convergence.In this work, we propose a joint cooperative end-to-end learning method for DCSS. It addresses many drawbacks with existing deep semantic segmentation learning; the proposed approach simultaneously learn both segmentation and classification; taking away the essential need of the pre-trained model for learning convergence. We present an improved inception based architecture with partial attention gating (PAG) over encoder information. The PAG also adds to achieve faster convergence and better accuracy for segmentation task. We will show the effectiveness of this learning on a diabetic retinopathy classification and segmentation dataset. 

\end{abstract}


%
\IEEEpeerreviewmaketitle

\section{Introduction}
Semantic segmentation task using deep learning is formulated as a pixel-wise classification problem \cite{segnet, pspnet, mscoco, gated, largekernel}. Generally, it's an encoder-decoder structure, one encoder to encode the features to lower dimensional features map and one decoder to remap the features map into several probabilistic maps (number of classes) of same width and height as input using bilinear interpolation or transposed convolutions \cite{zdeconv, deconv}. End-to-end learning of the encoder-decoder model best achieved using a pre-trained classification model for the encoder on the same domain. However, for a simple problem with low spatial resolutions input size and fewer number of classes pre-trained model isn't necessary, though it helps speed up the learning process. For complex problems with many numbers of classes and higher resolutions, learning step doesn't converge to the best local optimum without a pre-trained model; leading to inferior validation mean IOU(mean pixel intersection over union). For problems with completely unrelated domain such as medical images, satellite images etc. segmentation becomes two unrelated step; first, define and train a classification model and then use the same for the encoder to train the encoder-decoder model. \\
  How do we simplify the segmentation learning for any domains and complexity? Can we formulate the two-step processes into one step; at the same time learning both steps? 
We can formulate this as high-level multi-tasks learning concept to train the segmentation system. The idea of pure multi-tasks learning deviates from the main principle behind the simplification to learn one task from the feedback of another task. Here in our case segmentation is dependent on the pre-trained model, achieved using the classification step, which is completely independent of segmentation. In general, all tasks of a usual multi-tasks learning model can be learned independently; but here segmentation can't be learned independently. Another aspect of this attempt is that the classification step should be trainable on loose labels (Section~\ref{randomlabel}) of the dataset.
  
  Addressing the above concerns, in this work, we define a fully end-to-end learning system, that learns classification and segmentation jointly. We will show the model performance on a diabetic retinopathy (DR) classification and features segmentation dataset.\\
  
\subsection{Contribution}
\textbf{End-to-end joint segmentation and classification learning:} In contrast to the existing semantic segmentation models, the proposed learning system doesn't require a pre-trained model; it learns from the feedback of the classification node. 
  Advantages of a joint segmentation learning system?
  
    a. The scarcity of labeled segmentation data; this method can achieve better accuracy with the fewer number of segmentation ground-truths.
    
    b. Eliminating the need for pre-trained model requirements for convergence.
    
    c. Training complex model for any domains with larger input size.
    
    d. Learning a classification model as by-product alongside.

\textbf{Improved Residual Inception} We have proposed a new improved residual inception block. 

\textbf{Partial Attention:} A partial attention mechanism to use the mid-level encoder layer features for decoder layers.
 
\textbf{Training with partial random labels:} Learning method that can learn segmentation labeling while cooperatively training a classifier agent on partial random labels. 

\subsection{Related Work}
In this section, we will quickly go through some of the early and recent works on semantic segmentation using deep learning. Several recent works using encoder-decoder structure focus on improving the mean IOU for semantic segmentation benchmarks such as PASACAL-VOC\cite{voc_extra}, MS-COCO\cite{mscoco} and biomedical image datasets. One of the first success using the convolutional net for semantic segmentation was achieved in FCN\cite{fcn}; where fully connected layers were replaced using unit strided convolution and upsampling layers initialized with simple bilinear interpolation were used to reconstruct the feature map. In DeconvNet\cite{deconv}, a multilayer deconvolution and unpooling network were used to predict the segmentation masks. Segnet\cite{segnet} decoder uses encoder max-pooling indices to upsample low-resolution feature maps without learning the upsampling layers. Also, they have used convolution layer for each upsampling layer to get dense feature map. Recently, object detection and semantic segmentation were jointly learned in Mask R-CNN\cite{maskrcnn}, they have used FCN to predict segmentation mask for each region of interest. Also several other recent methods such as PSPNet\cite{pspnet}, RefineNet\cite{refinenet}, ENet\cite{enet}, Sharpmask\cite{sharpmask} etc. demonstrated the advantages of encoder-decoder model using convolution and transposed convolution for the task of semantic segmentation. In general, the predicted segmentation mask is coarse and requires post-processing to remove outliers. DenseCRF\cite{crf} is one effective post-processing layer for semantic segmentation \cite{crf1,crf2, crf3}, it refines the segmentation masks exploiting the pixel-level pairwise closeness. 

For biomedical image segmentation, deep learning based approach was proposed in Unet\cite{unet}. For diabetic retinopathy detection and microaneurysms segmentation \cite{drhaloi} using deep learning was proposed, this method adopted patch-wise classification for segmentation.

\section{Method}

\subsection{Cooperative Learning}
We formulate the segmentation learning as a cooperative learning problem, where two agents operate on two overlapping domains and interact with each other to achieve their respective goals. The terms overlapping domains implies that the datasets for classification and segmentation are sampled from the same distribution; it's not necessary to have an identical sample for both agents. One agent learns to generate segmentation masks for images with the help of the agent responsible for classification. Even though the classifier agent is capable of self-learning, the interaction with the segmentation agent results in improved generalizability. Each of the agents shares a chunk of parameters but uses separate optimizers to learn their parameters. The shared parameters of the agents are learned by both of the agents at a different rate; this is where the agents share knowledge and learn cooperatively. Learning rate for classifier agent is higher than that of segmentation agent as it possesses class specific discriminative knowledge about the domain. Outputs of the two agents Eq~\ref{eg:agents} for some input $x$ is calculated using the shared parameters $M$; $y_{common}$ denotes the output from the shared model part, $y_{clf}$ denotes the output of the classifier agent with private parameters $C$ and $y_{seg}$ denotes the output of the segmentation agent with private parameters $S$.

\begin{equation}
\label{eg:agents}
\begin{aligned}
y_{common} = M(x) \\
y_{clf} = C(M(x)) \\
y_{seg} = S(M(x)) 
\end{aligned}
\end{equation}
\begin{figure}[h]
  \centering
      \includegraphics[width=3.1in,height=1.7in]{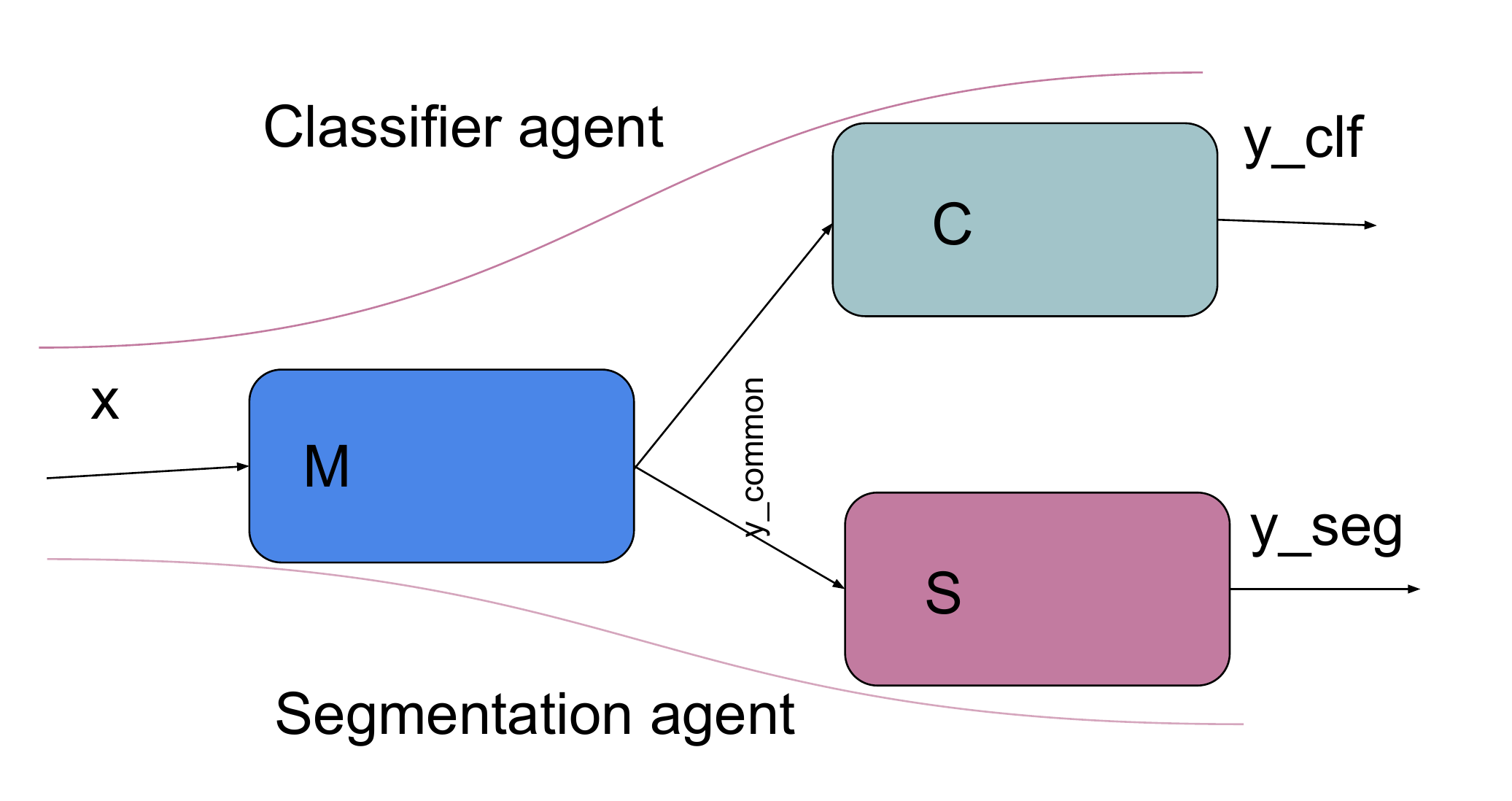}
\caption{Two agents and their shared parameters net M}
\label{fig:conv_pool}
\end{figure}
\begin{figure}[h]
  \centering
      \includegraphics[width=3.1in,height=2.7in]{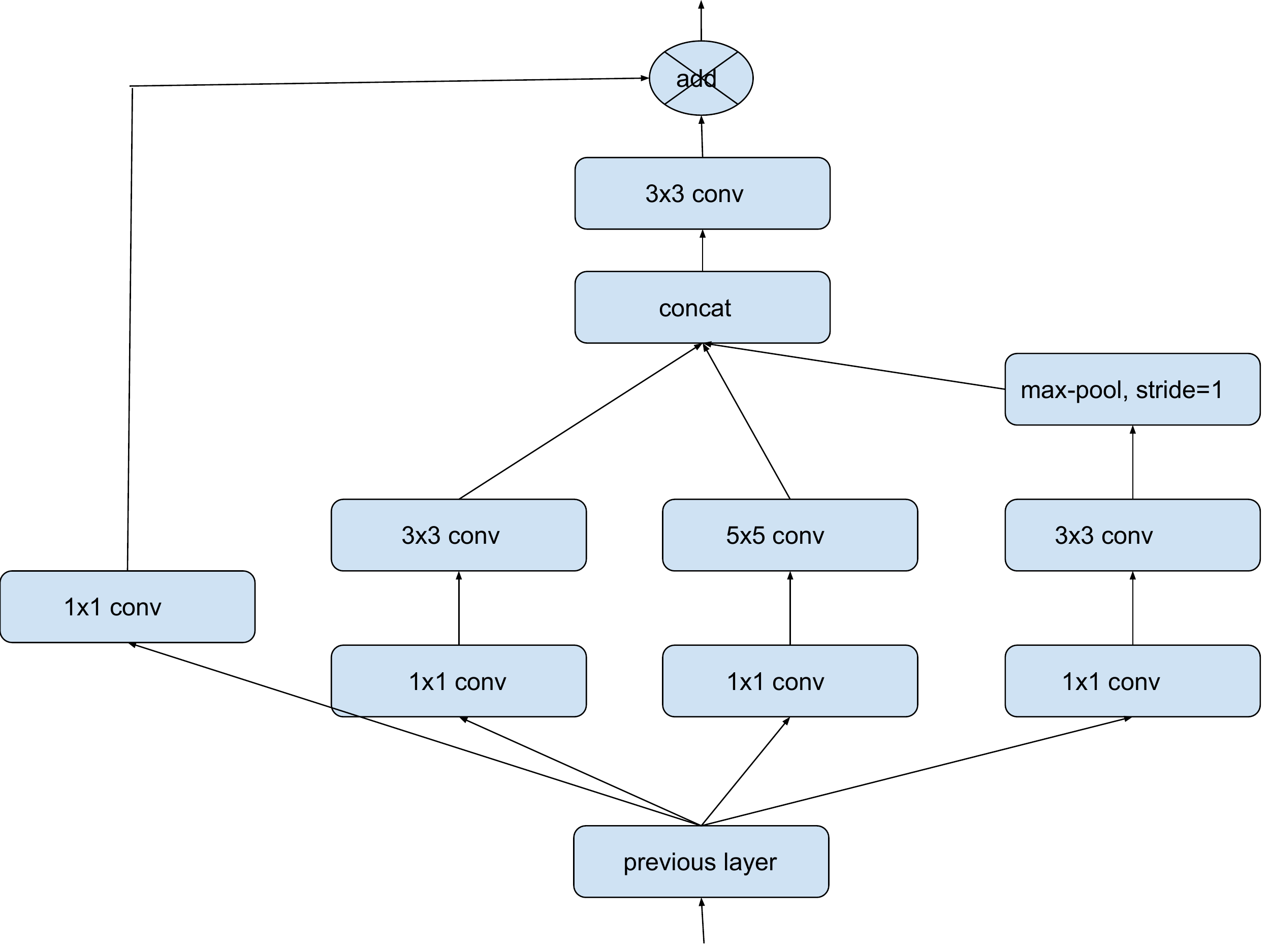}
\caption{Residual Inception}
\label{fig:incept}
\end{figure}
\begin{equation}
\label{eq:loss}
\begin{aligned}
J_{clf} = \min\limits_{\theta_{1} \cup \theta_{2}}  Cost(C(M(x)), y_{clf}^{label}) \\
J_{seg} = \min\limits_{\theta_{1} \cup \theta_{3}} Cost(S(M(x)), y_{seg}^{label}) \\
\theta_{1} \in M, \theta_{2} \in C, \theta_3 \in S
\end{aligned}
\end{equation}

Softmax cross-entropy cost Eq~\ref{eq:loss} was used to learn both agents. For the segmentation agent design and learning context, $M$ is refereed as the encoder and $S$ as the decoder. Fig~\ref{fig:conv_pool} shows the interaction between the two agents; sharing a common set of parameters $M$. 

\subsection{Network Architecture}
\begin{figure}[h]
  \centering
      \includegraphics[width=3.1in,height=2.7in]{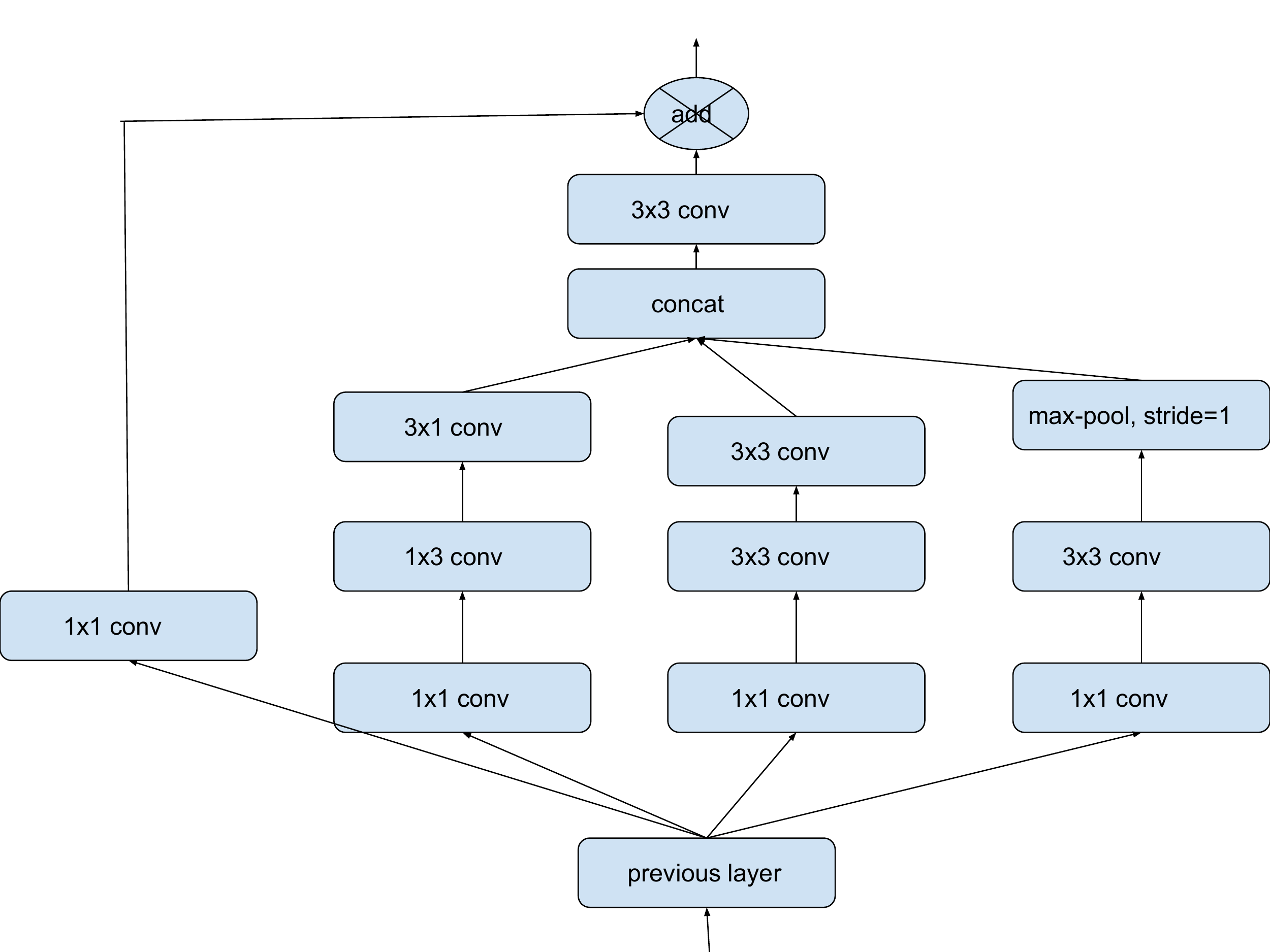}
\caption{Residual Inception with convolution factorization for the classifier node}
\label{fig:incept2}
\end{figure}

\textbf{Residual Inception module} We have used a modified version of the base inception block proposed in \cite{incept1}, where an additional convolution was used before max-pooling to ensure rich feature representation. We have added a residual connection from the block input to output as shown in Fig~\ref{fig:incept}. The concatenated features were convolved with another $3\times3$ convolution to reduce aliasing and added with $1\times1$ convolved input features; both operations have the same number of output filters. Apart from that, we have used a convolution factorization version Fig~\ref{fig:incept2} of the same module for the classifier node. The idea of convolution factorization was proposed in \cite{incept_google}, it reduces the number of parameters without effecting the generalization performance.

\begin{figure}[h]
  \centering
      \includegraphics[width=3.2in,height=2.3in]{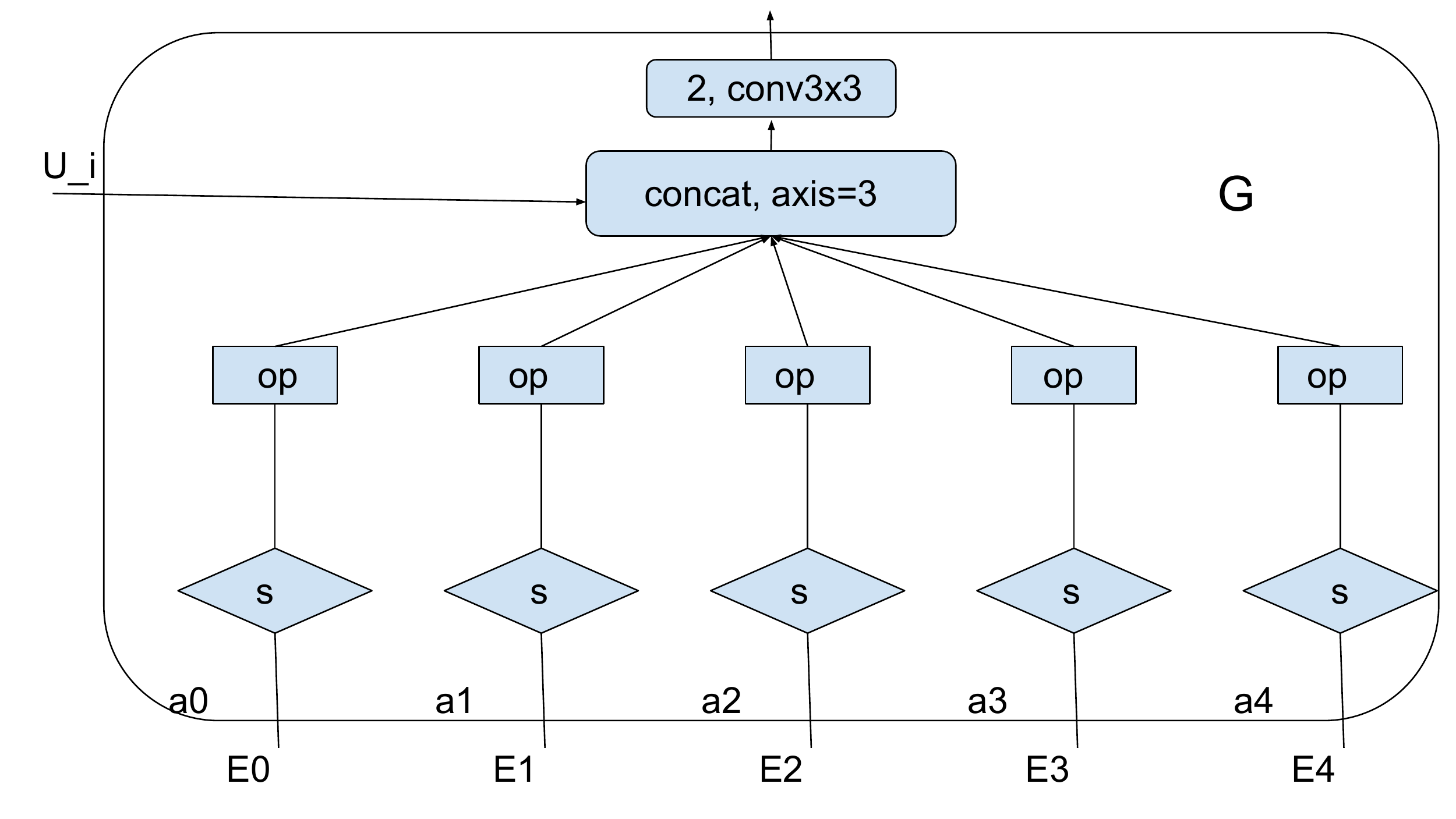}
\caption{Partial attention over encoder}
\label{fig:attention}
\end{figure}

\textbf{Partial attention}
We use partial attention over the selected encoder feature maps. Feeding the encoder information with upsampling boost the reconstructed feature map representation power. The attention mechanism here only focuses on the encoder feature maps with spatial resolutions equal or larger than that of the upsampling outputs. Max-pooling to reduce spatial resolution of encoder feature maps to the upsampling output size; max-pooling ensures maximally activated features. Fig~\ref{fig:attention} shows an overview of the partial attention mechanism used, where $E_i$ are the last feature maps (before reducing the feature map width and height for each block) of each convolution block of the encoder. The switch $s$ is on if the corresponding $a_i > 0$; $op$ is a max-pooling operation for all $a_i >1$ with stride $a_{i}$ and identity operation for $a_i=1$. 
\begin{equation}
a_i = \frac{min(height_{E_{i}}, width_{E_{i}})}{min(height_{U_{i}}, width_{U_{i}})}
\end{equation}

\textbf{Cooperative learning model}
follows the compositional design principles stated in \cite{deepcomp}; strided convolution is used to down-sample the encoder spatial resolution. To achieve richer feature representation, residual inception block as shown in Fig~\ref{fig:incept} is used for the encoder.  

\begin{figure*}
  \centering
      \includegraphics[width=5.3in,height=2.7in]{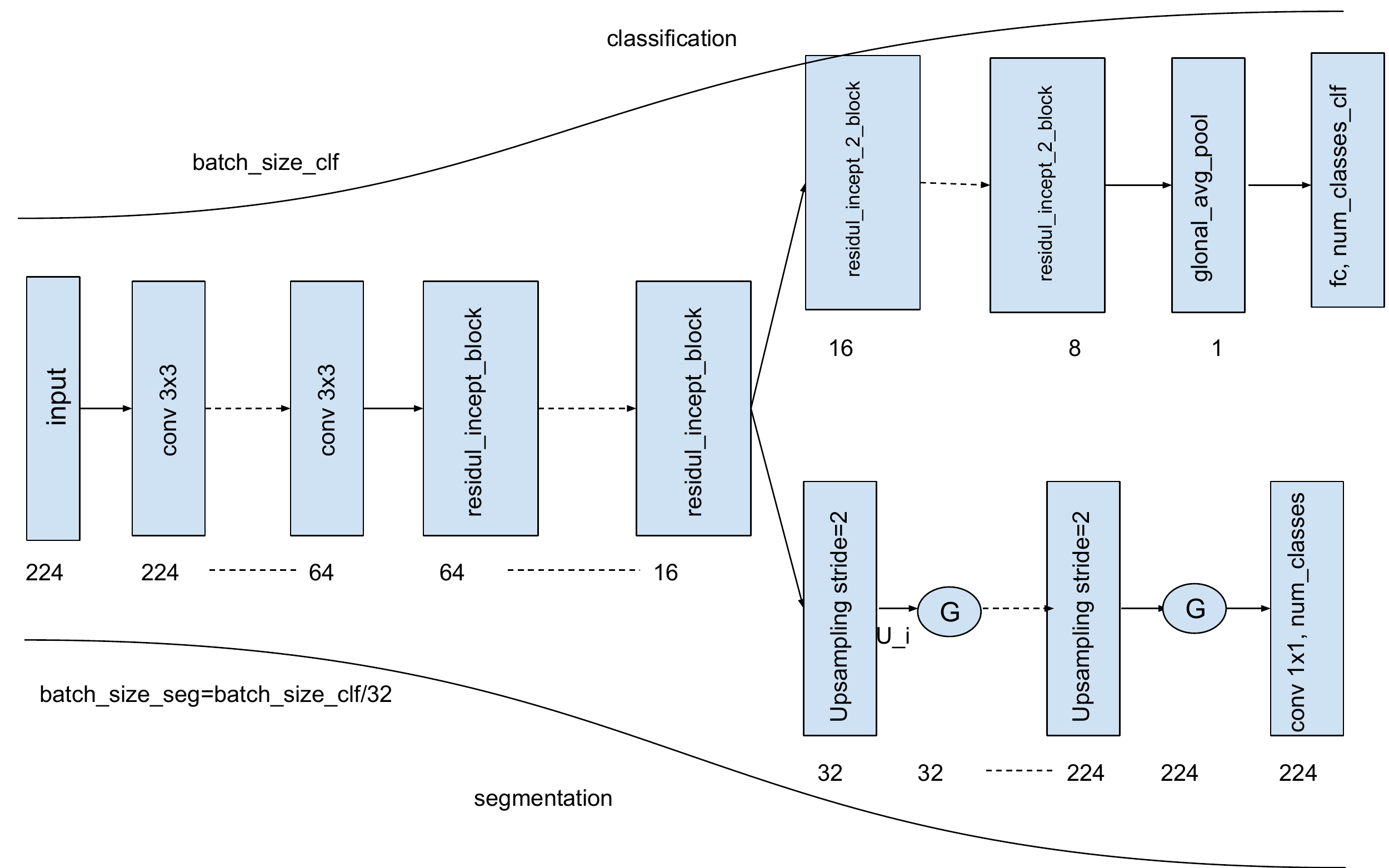}
\caption{Cooperative learning model with partial attention}
\label{fig:cooperative_model}
\end{figure*}

The encoder last layer output is used as the input of both the classifier and the segmentation agents. Classifier agent private node $C$ is designed using the residual factorized inception block as shown in Fig~\ref{fig:incept2}, followed by a global pool layer and a fully connected layer with a number of classes for the classifier. For each spatial dimension level of the encoder and the classifier private node, multiples cascaded residual inception blocks were used. Fig~\ref{fig:cooperative_model} shows the complete design of the joint segmentation and classification learning.

The decoder is designed using consecutive transposed convolutions with $2\times2$ stride and $2\times2$ kernel size followed by a partial attention block to incorporate encoder information for recovering object information. The partial attention block $G$ attend over few selective encoder feature maps with a spatial dimension larger or equal than that of the upsampled decoder output. Encoder feature maps learn input object information at multiple abstractions; the attention mechanism enables faster convergence of decoder parameters with improved generalization performance.

\subsection{Training on partial random labels}
\label{randomlabel}
The cooperative learning model can also be used when the number of labeled samples for the classifier agent is small. The classifier agent trained on partial random labels doesn't downgrade the segmentation agent performance. When we use around 20\% of labeled samples for classifier and around 80\% samples with random labels accuracy of classifier agent are lowered without effecting the performance of the segmentation agent. This observation indicates the ability of both of the agents to learn cooperatively. Also, this helps towards the goal of training on a dataset with the very low number of labeled images; especially relevant on medical image domains.

\section{Experiment Setup}
\textbf{Dataset:} We train the classification agent on the Eyepacs diabetic retinopathy classification dataset \cite{eyepacs} with 5 classes. The 5 classes of this dataset indicate DR severity level; level 0 for no DR, level 1 for mild DR, level 2 for moderate DR, level 3 for severe DR and 4 for proliferative DR. For training 25K images were used and validation performance was calculated on a set of 10k images. For testing, we have used 55K test set provided by eyepacs. The test set was labeled internally by ophthalmologists. Each image was resized to $224 \times 224 \times 3$. 
The segmentation agent is trained on a dataset prepared with the help of ophthalmologists; We selected a subset of around 5K images from Eyepacs dataset and annotated retinal pathological features with mask-based annotation. This dataset contains the annotation of around 5 pathological features (hard exudates, soft exudates, hemorrhage, microaneurysms, drusen) with one additional mask annotation for retinal artifacts. The set is split into training and validation set; 3.5K images for training and 1.5K images were used for validation. 

\textbf{Preprocessing:}
For both the classifier and the segmentation, $224\times224\times3$ is used as the input size. All input images were resized into $256\times256\times3$ and a randomly cropped patch of size $224\times224\times3$ is used as input. Extensive data augmentation is used such as random flip left/right, up/down and random padding was used for both classification and segmentation. Segmentation masks were also processed with same augmentation its corresponding image. Additionally, for classification additional augmentation such as for as changing the image pixel values randomly using hue, contrast and saturation were used. Also, each image is standardized by it's mean and dividing its standard deviation.   

\textbf{Framework:} We have used TEFLA\cite{tefla}, a python framework developed on the top of TENSORFLOW\cite{tf}, for all experiments described in this work.

\textbf{Training procedure:} Batch normalization\cite{bn} is used to reduce covariate shift and achieve faster convergence of both agents. Also, we have used Nesterov momentum optimizer with polynomial learning rate policy. For the classifier agent, the learning rate is 30 times than that of the segmentation agent. The batch size used for the classifier agent is 32 times than that of the segmentation agent. The use of higher batch size is to facilitate faster learning and information transfer from the classifier agent to the segmentation agent. We use lower learning rate for the segmentation agent to make sure that the information it passes to the classifier agent is minimal, but aid to generalization. 

\begin{figure*}
\centering
\begin{tabular}{cccc}
  \includegraphics[width=42mm]{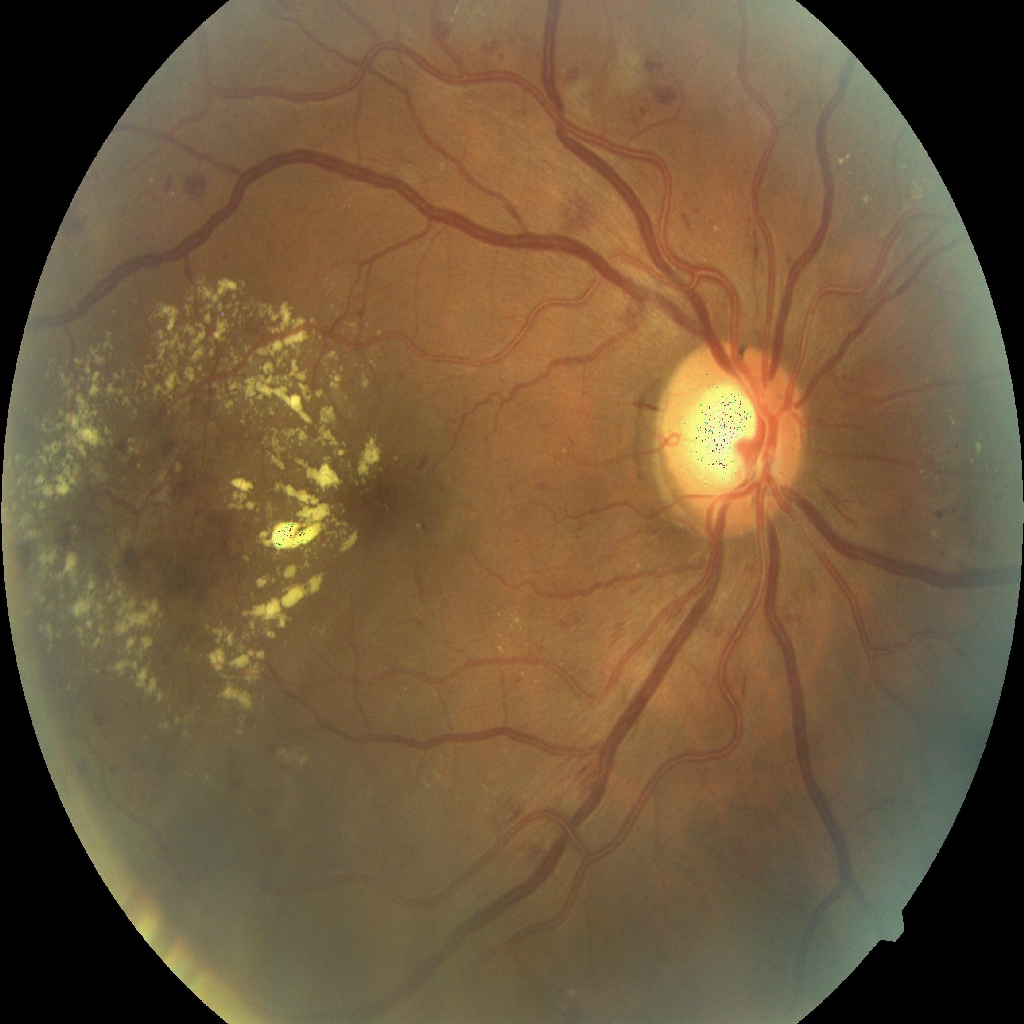} & \includegraphics[width=42mm]{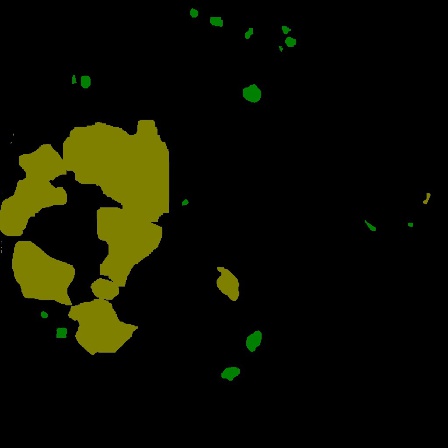} & \includegraphics[width=42mm]{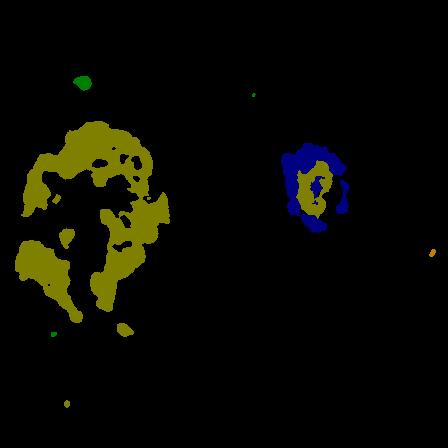} & \includegraphics[width=42mm]{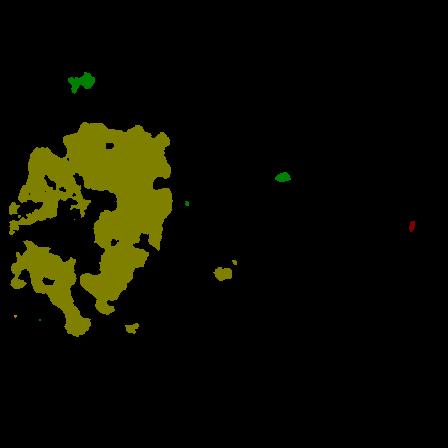}\\
   \includegraphics[width=42mm]{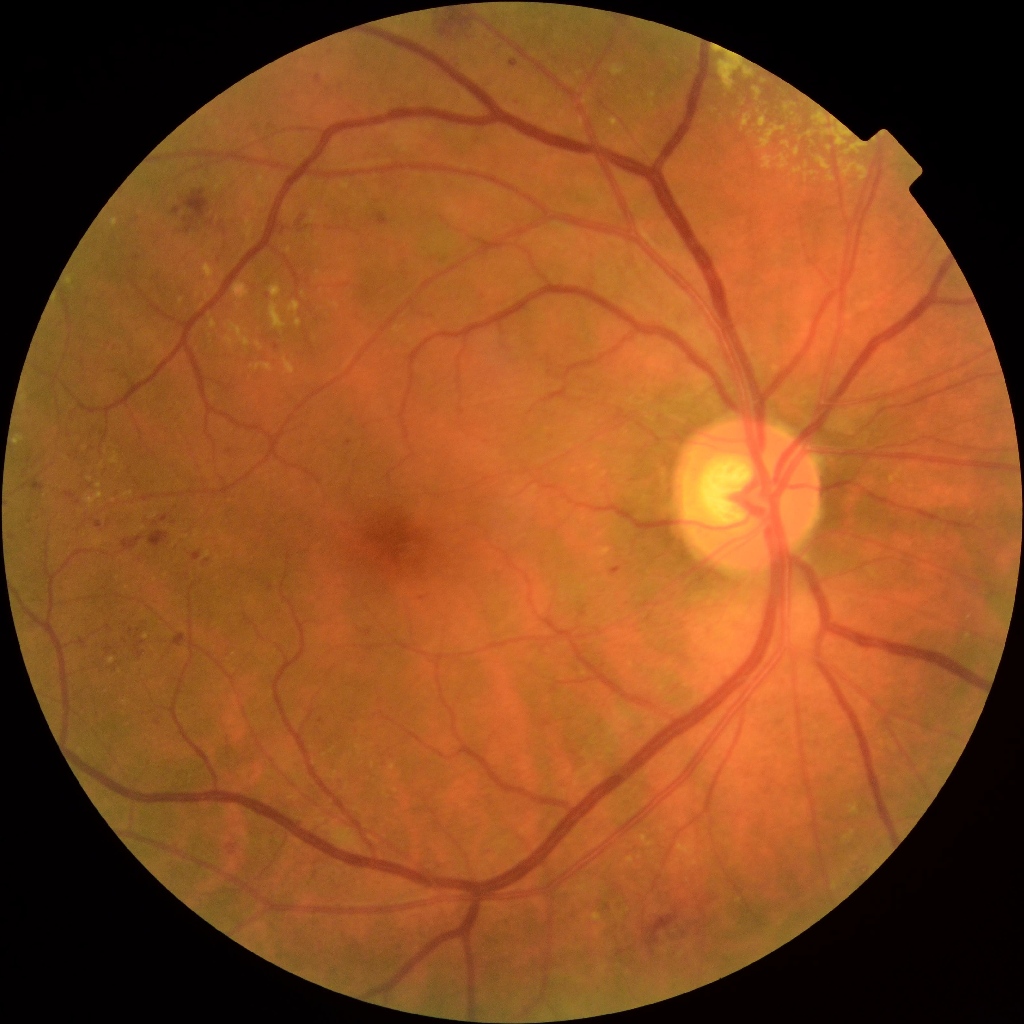} & \includegraphics[width=42mm]{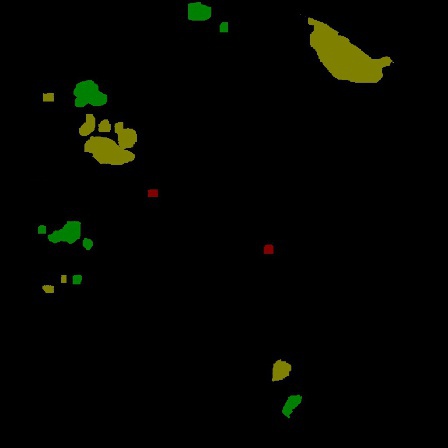} & \includegraphics[width=42mm]{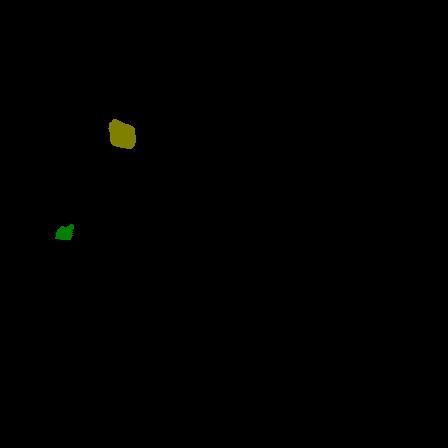} & \includegraphics[width=42mm]{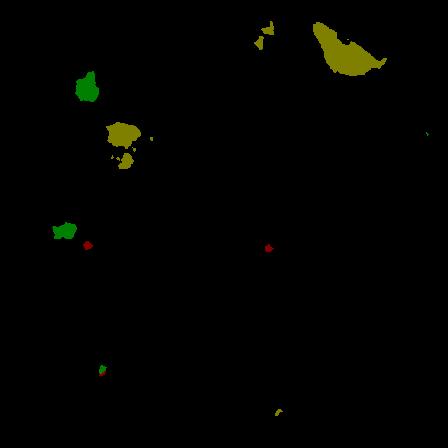}\\
 \includegraphics[width=42mm]{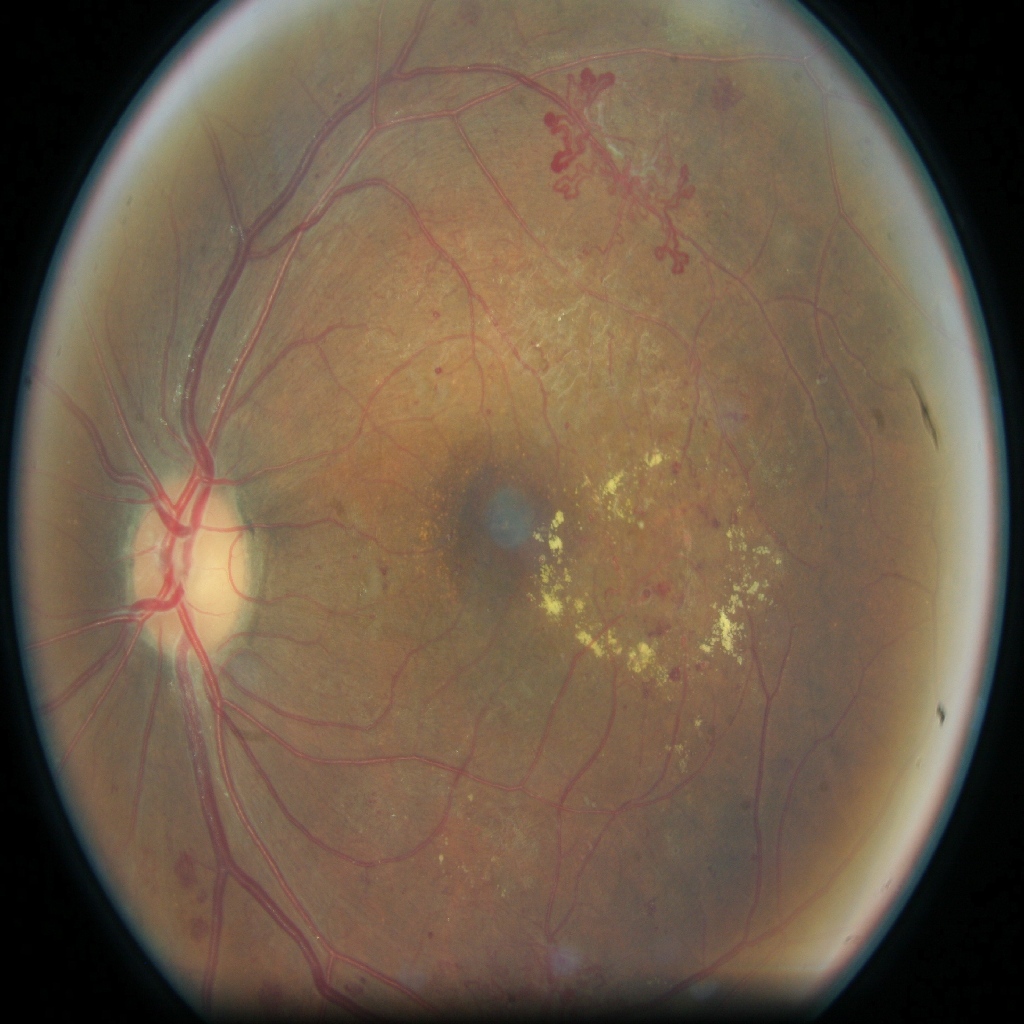} & \includegraphics[width=42mm]{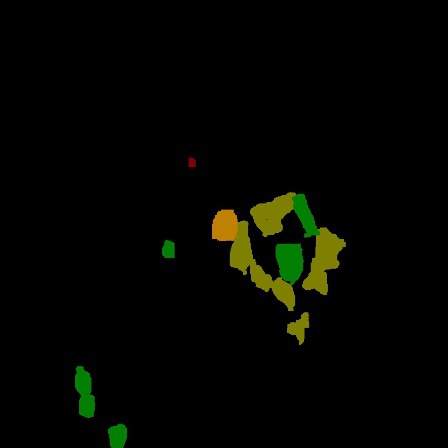} & \includegraphics[width=42mm]{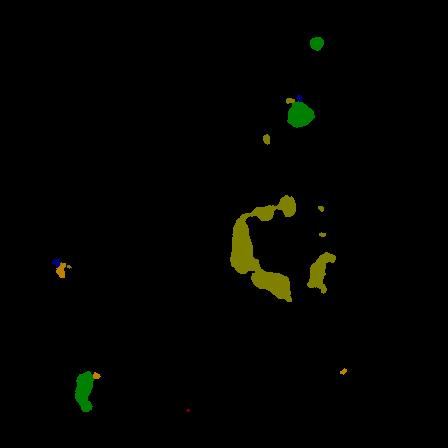} & \includegraphics[width=42mm]{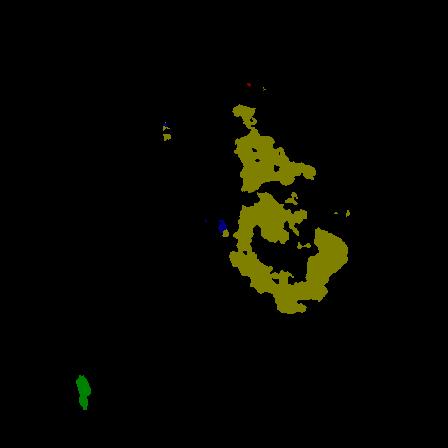}\\
Original Image & Ground truth & Unet\_inception & JointSeg[This work] \\
\end{tabular}
\caption{Results comparison of segmentations}. 
\label{fig:seg_result}
\end{figure*}

\subsection{Results Analysis}
We have evaluated out model performance on the eyepacs test set of 55k images. We have also trained VGG-19\cite{vgg} and inceptionv3\cite{incept_google} to compare classification agent performance with this proposed approach. Also for the segmentation agent performance comparisons we have trained FCN-8s\cite{fcn} and UNet\cite{unet} on the segmentation set. Here we use the improved inception block for UNet training; which is an improved version of the original UNet design.

Table~\ref{table:iou} shows the mean-IOU comparisons with the proposed approach and the existing algorithms. A significant performance gain is obtained with join segmentation learning. Here fcn-8s and Unet models parameters were initialized with a pre-trained model; trained on eyepacs classification set.

\begin{table}[h]
\begin{center}
  \begin{tabular}{ | p{3cm} | c | }
    \hline
    Method & mIOU \\ \hline
    FCN-8s & 55.3 \\ \hline
    UNet & 64.4 \\ \hline
    JointSeg [This work] & 66.1 \\ 
    \hline
  \end{tabular}
\end{center}
\caption{mIOU comparison on segmentation validation set using different methods}
\label{table:iou}
\end{table}
If we randomly initialize the FCN-8s and UNet model parameters; results become non-informatory as shown in Table~\ref{table:noinit}; both FCN-8s and UNet could not achieve convergence to an optimal local minimum. From Table~\ref{table:iou} and Table~\ref{table:noinit} we can understand the limitations and the importance of pre-trained model for traditional segmentation learning. But the joint learning achieves convergence to an optimal local minimum with random initializations. 
\begin{table}[h]
\begin{center}
  \begin{tabular}{ | p{3cm} | c | }
    \hline
    Method & mIOU \\ \hline
    fcn-8s & 15.3 \\ \hline
    Unet & 25.4 \\ \hline
    JointSeg [This work] & 66.1 \\ 
    \hline
  \end{tabular}
\end{center}
\caption{mIOU comparison with random parameters initializations}
\label{table:noinit}
\end{table}

Table~\ref{table:topk} shows the performance of the classifier agent with standard deep learning classifier model. 
\begin{table}[h]
\begin{center}
  \begin{tabular}{ | p{3cm} | c | c|}
    \hline
    Method & Top-1 (\%) & Top-2 (\%) \\ \hline
    VGG & 82.5 & 85.3 \\ \hline
    InceptionV3 & 84.3 & 87.2 \\ \hline
    JointSeg [This work] & 85.1 & 88.1\\ 
    \hline
  \end{tabular}
\end{center}
\caption{Accuracy comparison on classification test set using different methods}
\label{table:topk}
\end{table}

Table~\ref{table:bin} shows the classifier agent performance on a binary classification scenario. For a DR screening system to understand whether a person has DR, a binary label is enough. For this test, we combine all images with DR level higher than or equal to 1 as abnormal and DR level 0 as normal. The agent outperforms a normal human ophthalmologist for binary classification.

\begin{table}[h]
\begin{center}
  \begin{tabular}{ | p{3cm} | c |}
    \hline
    Method & Binary Accuracy (\%)  \\ \hline
    VGG & 89.5 \\ \hline
    InceptionV3 & 94.7 \\ \hline
    JointSeg [This work] & 95.5\\ \hline
    Human Performance & 95.1\\ 
    \hline
  \end{tabular}
\end{center}
\caption{Binary Accuracy comparison on classification test set using different methods}
\label{table:bin}
\end{table}

Figure~\ref{fig:seg_result} shows the qualitative performance of the segmentation agent with respect to the ground truths and modified UNet design. Joint Segmentation performs better than improved UNet design. Considering the complexity of this problem to identify very small lesions; the proposed architecture was able to learn meaningful features without a pre-trained model.

\section{Conclusion}
We have presented an end-to-end deep convolutional semantic segmentation learning method that achieves faster convergence and better accuracy with random parameters initializations. Also, the network learns both classification and segmentation on the same domain. We also showed the performance of our methods on biomedical image processing dataset eyepacs to detect and classify diabetic retinopathy. One of the important aspects of this work is that the classifier agent can be trained on lose/ partial random labels without degrading segmentation agent performance; this helps towards solving the label scarcity problems of medical image domain.


\section*{Acknowledgment}
We would like to thank the data team leads RajaRajalakshmi Kodhandapani and Ophthalmologists of Artelus; who have helped to prepare the segmentation dataset. Also, I would like to thank colleagues Anurag Sahay and Lalit Pant for their valuable advice during this work.



%

\end{document}